%% file: main.tex
\documentclass[runningheads]{llncs}
\usepackage{graphicx}
\usepackage{amsmath}
\usepackage{amssymb}
\usepackage{color}
\usepackage{bm}
\usepackage{textcomp} 
\usepackage{arydshln}
\usepackage{multirow}
\usepackage{mathtools}
\usepackage{microtype}
\usepackage{booktabs}
\usepackage{rotating}
\usepackage[]{algorithm2e}
\newcommand{\ra}[1]{\renewcommand{\arraystretch}{#1}}

\input{macro.tex}

\begin{document}
\pagestyle{headings}
\mainmatter
\title{{FLOT: Scene Flow on Point Clouds guided by Optimal Transport}}
\titlerunning{FLOT: Scene Flow by Optimal Transport}
\authorrunning{G.~Puy, A.~Boulch, and R.~Marlet}
\author{
Gilles Puy\inst{1} 
\and 
Alexandre Boulch\inst{1} 
\and 
Renaud Marlet\inst{1,2}}
\institute{
valeo.ai, Paris, France
\and
ENPC, Paris, France}
\maketitle

%
\begin{abstract}
\input{sections/abstract.tex}
\end{abstract}

%
\section{Introduction}
\input{sections/introduction.tex}

%
\section{Related Works}
\label{sec:related}
\input{sections/related.tex}

%
\section{Method}
\input{sections/method.tex}

%
\section{Experiments}
\label{sec:results}
\input{sections/results.tex}

%
\section{Conclusion}
\input{sections/conclusion.tex}

%
\appendix  
\input{sections/appendix.tex}

%
\bibliographystyle{splncs04}
\bibliography{biblio}

\end{document}

%% file: macro.tex
\newcommand{\ftp}{\ensuremath{\rm FT3D_{p}}}
\newcommand{\fts}{\ensuremath{\rm FT3D_{s}}}
\newcommand{\fto}{\ensuremath{\rm FT3D_{o}}}
\newcommand{\kittis}{\ensuremath{\rm KITTI_{s}}}
\newcommand{\kittio}{\ensuremath{\rm KITTI_{o}}}

\SetKwInput{KwInit}{Init}

\DeclareMathOperator*{\argmin}{argmin}
\DeclareMathOperator*{\argmax}{argmax}

\newcommand{\Rbb}{\ensuremath{\mathbb{R}}}

\newcommand{\inv}[1]{\ensuremath{\frac{1}{#1}}}

\renewcommand{\leq}{\ensuremath{\leqslant}}
\renewcommand{\geq}{\ensuremath{\geqslant}}

\newcommand{\adjoint}{\ensuremath{{\intercal}}}
\newcommand{\ma}[1]{\ensuremath{\mathsf{#1}}}
\renewcommand{\vec}[1]{\ensuremath{\bm{#1}}}
\newcommand{\diag}{\ensuremath{{\rm diag}}}

\newcommand{\norm}[1]{\ensuremath{\left\| #1\right\|}}

\newcommand{\set}[1]{\ensuremath{\mathcal{#1}}}

\renewcommand{\th}{\ensuremath{\text{th}}}
\newcommand{\eg}{\textit{e.g.}}
\newcommand{\ie}{\textit{i.e.}}

%% file: sections/abstract.tex
We propose and study a method called FLOT that estimates scene flow on point clouds. We start the design of FLOT by noticing that scene flow estimation on point clouds reduces to estimating a permutation matrix in a perfect world. Inspired by recent works on graph matching, we build a method to find these correspondences by borrowing tools from optimal transport. Then, we relax the transport constraints to take into account real-world imperfections. The transport cost between two points is given by the pairwise similarity between deep features extracted by a neural network trained under full supervision using synthetic datasets. Our main finding is that FLOT can perform as well as the best existing methods on synthetic and real-world datasets while requiring much less parameters and without using multiscale analysis. Our second finding is that,  on the training datasets considered, most of the performance can be explained by the learned transport cost. This yields a simpler method, FLOT$_0$, which is obtained using a particular choice of optimal transport parameters and performs nearly as well as FLOT.

%% file: sections/introduction.tex
Scene flow \cite{vedula_scene_flow} is the $3$D motion of points at the surface of objects in a scene. It is one of the low level information for scene understanding, which can be useful, \eg, in autonomous driving. Its estimation is a problem which has been studied for several years using different modalities as inputs such as colour images, with, \eg, variational approaches \cite{basha13}, \cite{wedel08} or methods using piecewise-constant priors \cite{ma_2019_CVPR}, \cite{menze_scene_flow_15}, \cite{vogel13}, or also using both colour and depth as modalities \cite{battrawy_stereo_lidar_flow}, \cite{hadfield_kinect}, \cite{shao_2018}. 

In this work,\footnote{The code is available at \url{https://github.com/valeoai/FLOT}.} 
we are interested in scene flow estimation on point clouds only using $3$D point coordinates as input. In this setting, \cite{dewan_2016} proposed a technique based on the minimisation of an objective function that favours closeness of matching points for accurate scene flow estimate and local smoothness of this estimate. In \cite{ushani17}, $2$D occupancy grids are constructed from the point clouds and given as input features to a learned background removal filter and a learned classifier that find matching grid cells. A minimisation problem using these grid matches is then proposed to compute a raw scene flow before a final refinement step. In \cite{ushani18a}, a similar strategy is proposed but the match between grid cells is done using deep features. In \cite{baur19}, \cite{zou_2019}, the point clouds are projected onto $2$D cylindrical maps and fed in a traditional CNN trained for scene flow estimation. In contrast, FLOT directly consumes point clouds by using convolutions defined for them. The closest related works are discussed in Section~\ref{sec:related}.

We split scene flow estimation into two successive steps. First, we find soft-correspondences between points of the input point clouds. Second, we exploit these correspondences to estimate the flow. Taking inspiration from recent works on graph matching that use optimal transport to match nodes/vertices in two different graphs \cite{maretic_got}, \cite{peyre_gw}, \cite{vayer_fgw}, we study the use of such tools for finding soft-correspondences between points. 

Our network takes as input two point clouds captured in the same scene at two consecutive instants $t$ and $t+1$. We extract deep features at each point using point cloud convolutions and use these features to compute a transport cost between the points at time $t$ and $t+1$. A small cost between two points indicates a likely correspondence between them. In the second step of the method, we exploit these soft-correspondences to obtain a first scene flow estimate by linear interpolation. This estimate is then refined using a residual network. The optimal transport and networks' parameters are learned by gradient descent under full supervision on synthetic datasets.

Our main contributions are: (\textit{a}) an optimal transport module for scene ﬂow estimation and the study of its performance; (\textit{b}) a lightweight architecture that can perform as well as the best existing methods on synthetic and real-world datasets with much less parameters and without using multiscale analyses; (\textit{c}) a simpler method FLOT$_0$ obtained for a particular choice of the OT parameters and which achieves competing results with respect to the state-of-the-art methods. We arrive at this simplified version by noticing that most of the performance in FLOT are explained by the learned transport cost. We also notice that the main module of FLOT$_0$ can be seen as an attention mechanism. Finally, we discuss, in the conclusion, some limitations of FLOT concerning the absence of explicit treatment of occlusions in the scene.

%% file: sections/related.tex
\textbf{Deep Scene Flow Estimation on Point Clouds.} 
In \cite{behl_pointflownet}, a deep network is trained end-to-end to estimate rigid motion of objects in LIDAR scans. The closest related works where no assumption of rigidity is made are \cite{gu_hplflownet}, \cite{liu_flownet3d}, \cite{wang_pccn}, \cite{wu_pointpwcnet}. In \cite{wang_pccn}, a parametric continuous convolution that operates on data lying on irregular structures is proposed and its efficiency is demonstrated on segmentation tasks and scene flow estimation.
The method \cite{liu_flownet3d} relies on PointNet++ \cite{qi_pointnet} and uses a new flow embedding layer that learns to mix the information of both point clouds to yield accurate flow estimates. In \cite{gu_hplflownet}, a technique to perform sparse convolutions on a permutohedral lattice is proposed. This method allows the processing of large point clouds. Furthermore, it is proposed to fuse the information of both point clouds at several scales, unlike in \cite{liu_flownet3d} where the information is fused once at a coarse scale. In contrast, our method fuse the information once at the finest scale. Let us highlight that our optimal transport module is independent of the type of point cloud convolution. We choose PointNet++ but other convolution could be used. In \cite{wu_pointpwcnet}, PWC-Net \cite{sun_pwcnet} is adapted to work on point clouds. The flow is estimated in a coarse-to-fine scale fashion showing improvement over the previous method. Finally, let us mention that recent works \cite{mittal_selfflow}, \cite{wu_pointpwcnet} address this topic using self-supervision. We however restrict ourselves to full supervision in this work.

\textbf{Graph Matching by Optimal Transport.} Our method is inspired by recent works on graphs comparison using optimal transport. In \cite{maretic_got}, the graph Laplacian is used to map a graph to a multidimensional Gaussian distribution that represents the graph structure. The Wasserstein distance between these distributions is then used as a measure of graph similarity and permits one to match nodes between graphs. In \cite{nikolentzos_graph}, each graph is represented as a bag-of-vectors (one vector per node) and the measure of similarity is the Wasserstein distance between these sets. In \cite{peyre_gw}, a method building upon the Gromov-Wasserstein distance between metric-measure spaces \cite{memoli_gw} is proposed to compare similarity matrices. This method can be used to compare two graphs by, \eg, representing each of them with a matrix containing the geodesic distances between all pairs of nodes. In \cite{vayer_fgw}, it is proposed to compare graphs by fusing the Gromov-Wassertsein distance with the Wasserstein distance. The former is used to compare the graph structures while the latter is used to take into account node features. In our work, we use the latter distance. A graph is constructed for each point cloud by connecting each point to its nearest neighbours. We then propose a method to train a network that extract deep features for each point and use these features to match points between point clouds in our optimal transport module.

\textbf{Algorithm Unrolling.} Our method is based on the algorithm unrolling technique which consists in taking an iterative algorithm, unrolling a fixed number of its iterations, and replacing part of the matrix multiplications/convolutions in these unrolled iterations by new ones trained specifically for the task to achieve. Several works build on this technique, such as \cite{gregor_lista}, \cite{mardani_prox}, \cite{metzler_amp}, \cite{mousavi_invert} 
to solve linear inverse problems, or 
\cite{chen_diffusion}, \cite{liu_rare}, \cite{meinhart_prox}, \cite{wang_2016} 
in for image denoising (where the denoiser is sometimes used to solve yet another inverse problem). In this work, we unroll few iterations of the Sinkhorn algorithm and train the cost matrix involved in it. This matrix is trained so that the resulting transport plan provides a good scene flow estimate. Let us mention that this algorithm is also unrolled, \eg, in \cite{genevay_generative} to train a deep generative network, and in \cite{sarlin_feat_matching} for image feature assignments.

%% file: sections/method.tex
\subsection{Step 1: Finding Soft-Correspondences between Points}

Let $\vec{p}, \, \vec{q} \in \Rbb^{n \times 3}$ be two point clouds of the same scene at two consecutive instants $t$ and $t+1$. The vectors $\vec{p}_i, \vec{q}_j \in \Rbb^3$ are the $xyz$ coordinates of the $i^\th$ and $j^\th$ points of $\vec{p}$ and $\vec{q}$, respectively. The scene flow estimation problem on point clouds consists in estimating the scene flow $\vec{f} \in \Rbb^{n \times 3}$ where $\vec{f}_i \in \Rbb^3$ is the translation of $\vec{p}_i$ from $t$ to $t+1$.

\subsubsection{Perfect World.}

\begin{figure}[t]
\centering
\includegraphics[width=\linewidth]{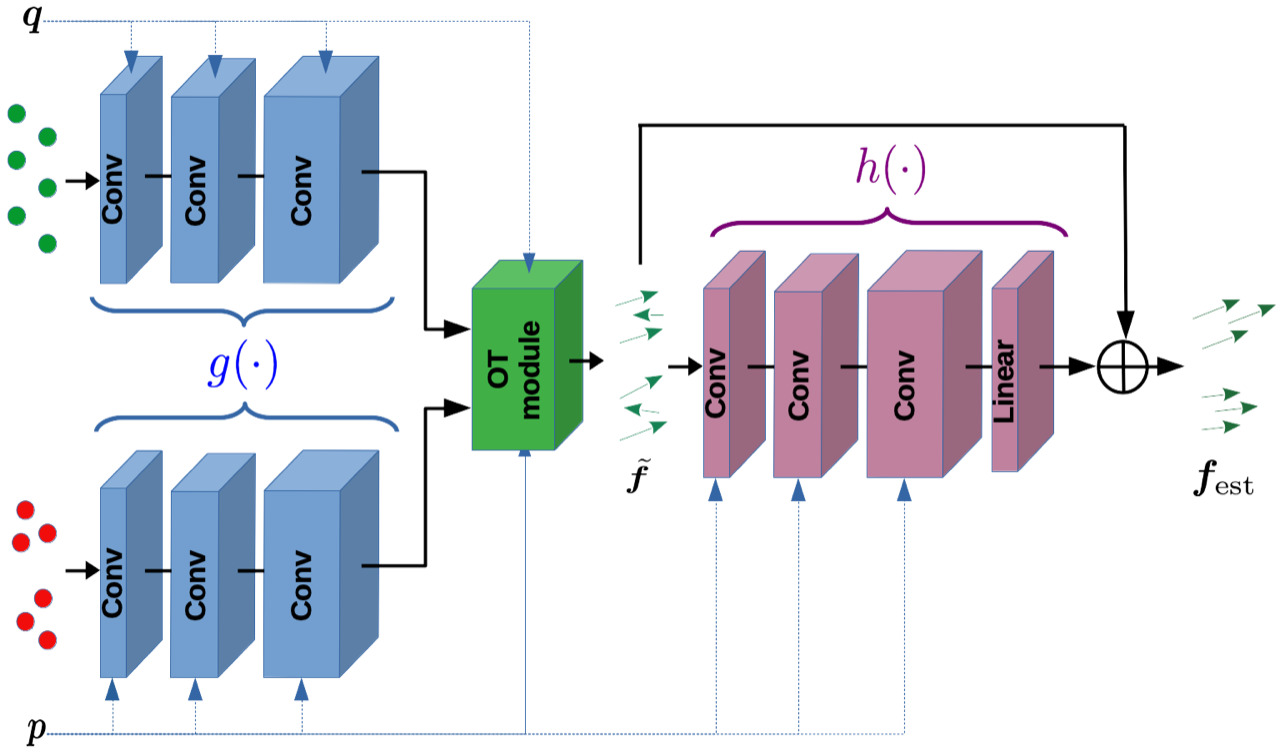}
\caption{\label{fig:network}The point clouds $\vec{p}$ and $\vec{q}$ go through $g$ which outputs a feature for each input point. These features (black arrows) go in our proposed OT module where they are used to compute the pairwise similarities between each pair of points $(\vec{p}_i, \vec{q}_j)$. The output of the OT module is a transport plan which informs us on the correspondences between the points of $\vec{p}$ and $\vec{q}$. This information permits us to compute a first scene flow estimate $\tilde{\vec{f}}$, which is refined by $h$ to obtain $\vec{f}_{\rm est}$. The convolution layers (conv) are based on PointNet++ \cite{qi_pointnet} but the OT module could accept the output of any other point cloud convolution. The dashed-blue arrows indicate that the point coordinates are passed to each layer to be able to compute convolutions on points.}
\end{figure}

We construct FLOT starting in the perfect world where $\vec{p} + \vec{f} = \ma{P} \, \vec{q}$, with $\ma{P} \in \{0, 1\}^{n \times n}$ a permutation matrix. The role of FLOT is to estimate the permutation matrix $\ma{P}$ without the knowledge of $\vec{f}$. In order to do so, we use tools from optimal transport. We interpret the motion of the points $\vec{p}_i$ as a displacement of mass between time $t$ and $t+1$. Each point in the first point cloud $\vec{p}$ is attributed a mass which we fix to $n^{-1}$. Each point $\vec{q}_j$ then receives the mass $n^{-1}$ from $\vec{p}_i$ if $\vec{p}_i + \vec{f}_i = \vec{q}_j$, or, equivalently, if $\ma{P}_{ij} = 1$. We propose to estimate the permutation matrix $\ma{P}$ by computing a transport plan $\ma{T} \in \Rbb_+^{n \times n}$ from $\vec{p}$ to $\vec{q}$ which satisfies
\begin{align}
\label{eq:optimal_transport}
\ma{T} \in 
\argmin_{\ma{U} \in \Rbb_+^{n \times n}} 
\sum_{i,j = 1}^n \ma{C}_{ij} \ma{U}_{ij} 
\quad \text{ subject to } \quad
\ma{U} \vec{1} = \vec{1}  n^{-1} 
\; \text{ and } \; 
\ma{U}^\adjoint \vec{1} = \vec{1}  n^{-1},
\end{align}
where $\vec{1} \in \Rbb^{n}$ is the vector with all entries equal to $1$, and $\ma{C}_{ij} \geq 0$ is the displacement cost from point $\vec{p}_i$ to point $\vec{q}_j$ \cite{peyre_cot}. Each scalar entry $\ma{T}_{ij} \geq 0$ of the transport plan $\ma{T}$ represents the mass that is transported from $\vec{p}_i$ to $\vec{q}_j$. 

The first constraint in \eqref{eq:optimal_transport} imposes that the mass of each point $\vec{p}_i$ is entirely distributed over some of the points in $\vec{q}$. The second constraint imposes that each points $\vec{q}_j$ receives exactly a mass $n^{-1}$ from some of the points $\vec{p}$. No mass is lost during the transfer. Note that in the hypothetical case where the cost matrix $\ma{C}$ would contain one zero entry per line and per column then the transport plan is null everywhere except on these entries and the mass constraints are immediately satisfied via a simple scaling of the transport plan. In this hypothetical situation, the mass constraints would be redundant for our application as it would have been enough to find the zero entries of $\ma{C}$ to estimate $\ma{P}$. It is important to note the mass constraints play a role in the more realistic situation where ``ambiguities'' are present in $\ma{C}$ by ensuring that each point gives/receives a mass $n^{-1}$ and that each point in $\vec{p}$ has a least one corresponding point in $\vec{q}$ and vice-versa.

We note that $n^{-1} \ma{P}$ satisfies the optimal transport constraints. We need now to construct $\ma{C}$ so that $\ma{T} = n^{-1} \ma{P}$.

\subsubsection{Real World and Fast Estimation of \ma{T}.}

\begin{algorithm}[t]
\KwIn{cost matrix $\ma{C}$; parameters $K, \lambda, \epsilon > 0$.}
\KwOut{transport plan $\ma{T}$.}
$\bm{a} \leftarrow \vec{1} n^{-1}$\; 
$\ma{U} \leftarrow \exp(- \, \ma{C} / \epsilon)$\;
\For{$k=1, \ldots, K$}{
	$\vec{b} \leftarrow [(\vec{1} n^{-1}) \oslash (\ma{U}^\adjoint \vec{a})]^{\lambda / (\lambda + \epsilon)}$ \;
	$\vec{a} \leftarrow [(\vec{1} n^{-1}) \oslash (\ma{U} \; \vec{b})]^{\lambda / (\lambda + \epsilon)}$ \;
}
$\ma{T} \leftarrow \diag(\vec{a}) \; \ma{U} \;\diag(\vec{b})$ \;
\caption{Optimal transport module. The symbol $\oslash$ denote the element-wise division and multiplication, respectively.}
\label{alg:sinkhorn}
\end{algorithm}

In the real world, the equality $\vec{p} + \vec{f} = \ma{P} \, \vec{q}$ does not hold because the surfaces are not sampled at the same physical locations at $t$ and $t+1$ and because objects can (dis)appear due to occlusions. A consequence of these imperfections is that the mass preservation in \eqref{eq:optimal_transport} does not hold exactly: mass can (dis)appear. One solution to circumvent this issue is to relax the constraints in \eqref{eq:optimal_transport}. Instead of solving \eqref{eq:optimal_transport}, we propose to solve
\begin{align}
\label{eq:optimal_transport_reg}
\min_{\ma{U} \in \Rbb_+^{n \times n}} 
\left[
\sum_{i,j = 1}^n 
\ma{C}_{ij} \ma{U}_{ij} 
+ \epsilon \, 
    \ma{U}_{ij} \left(\log{\ma{U}_{ij}} - 1\right) 
\right] 
+  \lambda \, 
    {\rm KL} \left(\ma{U} \vec{1}, \, \frac{\vec{1}}{n} \right) 
+ \lambda \, 
    {\rm KL} \left(\ma{U}^\adjoint \vec{1}, \, \frac{\vec{1}}{n} \right),
\end{align}
where $\epsilon, \lambda \geq 0$, and ${\rm KL}$ denotes the ${\rm KL}$-divergence. The term $\ma{U}_{ij} (\log{\ma{U}_{ij}} - 1)$ in \eqref{eq:optimal_transport_reg} is an entropic regularisation on the transport plan. Its main purpose, in our case, is to allow the use of an efficient algorithm to estimate the transport plan: the Sinkhorn algorithm \cite{cuturi_sinkhorn}. The version of this algorithm for the optimal transport problem \eqref{eq:optimal_transport_reg} is derived in \cite{chizat_unbalancedot} and is presented in Alg.~\ref{alg:sinkhorn}. The parameter $\epsilon$ controls the amount of entropic regularisation. The smaller $\epsilon$ is, the sparser the transport plan is, hence finding sparse correspondences between $\vec{p}$ and $\vec{q}$. The regularisation parameter $\lambda$ adjust how much the transported mass deviates from the uniform distribution, allowing mass variation. One could let $\lambda \rightarrow +\infty$ to impose strict mass preservation. 

Note that the mass regularisation is controlled by the power $\lambda / (\lambda + \epsilon)$ in Alg.~\ref{alg:sinkhorn}. This power tends to $1$ when $\lambda \rightarrow +\infty$ to impose strict mass preservation and reaches $0$ in absence of any regularisation. Instead of fixing the parameters $\epsilon, \lambda$ in advance, we let these parameters free and learn them by gradient descent along with the other networks' parameters.

We would like to recall that, in the perfect world, it is not necessary for the power $\lambda / (\lambda + \epsilon)$ to reach $1$ to yield accurate results as the final quality is also driven by the quality of $\ma{C}$. In a perfect situation where the cost would be perfectly trained with a bijective mapping already encoded in $\ma{C}$ by its zero entries, then any amount of mass regularisation is sufficient to reach accurate results. This follows from our remark at the end of the previous subsection but also from the discussion in the subsection below on the role of $\ma{C}$ and the mass regularisation. In a real situation, the cost is not perfectly trained and we expect the power $\lambda / (\lambda + \epsilon)$ to vary in the range of $(0, 1)$ after training, reaching values closer to $1$ when trained in a perfect world setting and closer to $0$ in presence of occlusions.

\subsubsection{Learning the Transport Cost.} 

An essential ingredient in \eqref{eq:optimal_transport_reg} is the cost $\ma{C} \in \Rbb^{n \times n}$ where each entry $\ma{C}_{ij}$ encodes the similarity between $\vec{p}_i$ to point $\vec{q}_j$. An obvious choice could be to take the Euclidean distance between each pair of points $(\vec{p}_i , \vec{q}_j)$, \ie, $\ma{C}_{ij} = \| \vec{p}_i - \vec{q}_j \|_2$, but this choice does not yield accurate results. In this work, we propose to learn the displacement costs by training a deep neural network $g: \mathbb{R}^{n \times 3} \rightarrow \mathbb{R}^{n \times c}$ that takes as input a point cloud and output a feature of size $c$ for each input point. The entries of the cost matrix are then defined using the cosine distance between the features $g(\vec{p})_i, g(\vec{q})_j \in \Rbb^{c}$ at points $\vec{p}_i$ and $\vec{q}_j$, respectively:
\begin{align}
\label{eq:cost}
\ma{C}_{ij} = 
\left(
1 - 
\frac
{g(\vec{p})_i^{\adjoint} \; g(\vec{q})_j}
{\norm{g(\vec{p})_i}_2 \norm{g(\vec{q})_j}_2} 
\right) 
\, \cdot \, 
i_{\norm{\cdot}_2 \leq d_{\rm max}} \, \left( \vec{p}_i - \vec{q}_j \right).
\end{align}
The more similar the features $g(\vec{p})_i$ and $g(\vec{q})_j$ are, the less the cost of transporting a unit mass from $\vec{p}_i$ to $\vec{q}_j$ is. The indicator function
\begin{align}
i_{\norm{\cdot}_2 \leq d_{\rm max}} \, \left( \vec{p}_i - \vec{q}_j \right)
=
\left\{
\begin{array}{l}
1 \text{ if } \norm{\vec{p}_i - \vec{q}_j}_2 \leq d_{\rm max}, \\
+\infty \text{ otherwise},
\end{array}
\right.
\end{align}
is used to prevent the algorithm to find correspondences between points too far away from each other. We set $d_{\rm max} = 10$ m.

In order to train the network $g$, we adopt the same strategy as, \eg, in \cite{genevay_generative} to train generative models or in \cite{sarlin_feat_matching} for matching image features. The strategy consists in unrolling $K$ iterations of Alg.~\ref{alg:sinkhorn}. This unrolled iterations constitute our OT module in Fig.~\ref{fig:network}. One can remark that the gradients can backpropagate through each step of this module and allow us to train $g$.

\subsubsection{On the Role of $\ma{C}$ and of the Mass Regularisation.} 
We gather in this paragraph the earlier discussions on the role of $\ma{C}$ and the mass regularisation. For the sake of the explanation, we come back in the perfect-world setting and consider \eqref{eq:optimal_transport}. In this ideal situation, one could further dream that it is possible to train $g$ perfectly such that $\ma{C}_{ij}$ is null for matching points, \ie, when $\ma{P}_{ij} = 1$, and strictly positive otherwise. The transport plan would then satisfy $\ma{T} = n^{-1} \ma{P}$ with a null transport cost. However, one should note that the solution $\ma{T}$ would entirely be encoded in $\ma{C}$ up to a global scaling factor: the non-zero entries of $\ma{T}$ are at the zero entries of $\ma{C}$. In that case, the mass transport constraints only adjust the scale of the entries in $\ma{T}$. Such a perfect scenario is unlikely to occur but these considerations highlight that the cost matrix $\ma{C}$ could be exploited alone and could maybe be sufficient to find the appropriate correspondences between $\vec{p}$ and $\vec{q}$ for scene flow estimation. The mass transport regularisation plays a role in the more realistic case where ambiguities appears in $\ma{C}$. The regularisation enforces, whatever the quality of $\ma{C}$ and with a ``strength'' controlled by $\lambda$, that the mass is distributed as uniformly as possible over all points. This avoids that some points in $\vec{p}$ are left with no matching point in $\vec{q}$, and vice-versa.

\subsubsection{FLOT$_0$.} FLOT$_0$ is a version of FLOT where only the cost matrix $\ma{C}$ is exploited to find correspondences between $\vec{p}$ and $\vec{q}$. This method is obtained when removing the mass transport regularisation in \eqref{eq:optimal_transport_reg}, \ie, by setting $\lambda = 0$. In this limit, the ``transport plan'' satisfies
\begin{align}
\ma{T} = \exp(- \, \ma{C} / \epsilon).
\end{align}
$\ma{T}$ is then used in the rest of the method as if it was the output of Alg.~\ref{alg:sinkhorn}.

\subsection{Step 2: Flow Estimation from Soft-Correspondences}

We obtained, in the previous step, a transport plan $\ma{T}$ that gives correspondences between the points of $\vec{p}, \vec{q}$. Our goal now is to exploit these correspondences to estimate the flow. As before, it is convenient to start in the perfect world and consider \eqref{eq:optimal_transport}. In this setting, we have seen that $\vec{f} = \ma{P} \vec{q} - \vec{p}$ and that, if $g$ is well trained, we expect $n^{-1} \ma{P} = \ma{T}$. Therefore, an obvious estimate of the flow is 
\begin{align}
\tilde{\vec{f}}_i 
= 
\sum_{j=1}^{n} \ma{P}_{ij} \; \vec{q}_{j} - \vec{p}_{i}
=
\frac{1}{n^{-1}} \sum_{j=1}^{n} \ma{T}_{ij} \; \vec{q}_{j} - \vec{p}_{i}
=
\label{eq:interpolation}
\frac{\sum_{j=1}^{n} \ma{T}_{ij} \; \vec{q}_{j}}{\sum_{j=1}^{n} \ma{T}_{ij}} 
- \vec{p}_{i},
\end{align}
where we exploited the fact that $\sum_{j=1}^{n} \ma{T}_{ij} = n^{-1}$ in the last equality.

In the real world, the first equality in \eqref{eq:interpolation} does not hold. Yet, the last expression in \eqref{eq:interpolation} remains a sensible first estimation of the flow. Indeed, this computation is equivalent to computing, for each point $\vec{p}_{i}$, a corresponding virtual point that is a barycentre of some points in $\vec{q}$. The larger the transported mass $\ma{T}_{ij}$ from $\vec{p}_{i}$ to $\vec{q}_{j}$ is, the larger the contribution of $\vec{q}_{j}$ to this virtual point is. The difference between this virtual point and $\vec{p}_{i}$ gives an estimate of the flow $\vec{f}_i$. This virtual point is a ``guess'' on the location of $\vec{p}_i + \vec{f}_i$ made knowing where the mass from $\vec{p}_i$ is transported in $\vec{q}$.

However, we remark that the flow $\tilde{\vec{f}}$ estimated in \eqref{eq:interpolation} is, necessarily, still imperfect as it is highly likely that some points in $\vec{p} + \vec{f}$ cannot be expressed as barycentres of the found corresponding points $\vec{q}$. Indeed, some portion of objects visible in $\vec{p}$ might not visible any more in $\vec{q}$ due to the finite resolution in point cloud sampling. The flow in these missing regions cannot be reconstructed from $\vec{q}$ but has to be reconstructed using structural information available in $\vec{p}$, relying on neighbouring information from the well sampled regions. Therefore, we refine the flow using a residual network:
\begin{align}
\vec{f}_{\rm est} = \tilde{\vec{f}} + h(\tilde{\vec{f}}),
\end{align}
where $h: \Rbb^{n \times 3} \rightarrow \Rbb^{n \times c}$ takes as inputs the estimated flow $\tilde{\vec{f}}$ and uses convolutions defined on the point cloud $\vec{p}$. 

Let us finally conclude this section by highlighting the fact that, in the case of FLOT$_0$, \eqref{eq:interpolation} simplifies to
\begin{align}
\tilde{\vec{f}}_i 
= 
\frac
{\sum_{j=1}^{n} \exp(- \, \ma{C}_{ij} / \epsilon) \, (\vec{q}_{j} - \vec{p}_{i})}
{\sum_{j=1}^{n} \exp(- \, \ma{C}_{ij} / \epsilon)}.  
\end{align}
On can remark that the OT module essentially reduces to an attention mechanism \cite{attention_2017} in that case. The attention mechanism is thus a particular case of FLOT where the entropic regularisation $\epsilon$ plays the role of the softmax temperature. Let us mention that similar attention layers haved been showed effective in related problems such as rigid registration \cite{wang_cycle,dcp_19,prnet_19}.

\subsection{Training}

The network's parameters, denoted by $\theta$, and $\epsilon, \gamma$ are trained jointly under full supervision on annotated synthetic datasets of size $L$. Note that to enforce positivity of $\epsilon, \gamma$, we learn their $\log$ values. A constant offset of $0.03$ is applied to $\epsilon$ to avoid numerical instabilities in the exponential function during training.

The sole training loss is the $\ell_1$-norm between the ground truth flow $\vec{f}$ and the estimated flow $\vec{f}_{\rm est}$:
\begin{align}
\label{eq:loss}
\min_{\theta} \inv{3L} \sum_{l=1}^{L} \norm{ \ma{M}^{(l)} \, (\vec{f}_{\rm est}^{(\ell)} -\vec{f}^{(\ell)})}_1,
\end{align}
where $\ma{M}^{(l)} \in \Rbb^{n \times n}$ is a diagonal matrix encoding an annotated mask used to remove points where the flow is occluded.

We use a batchsize of $4$ at $n=2048$ and a batchsize of $1$ at $n=8192$ using Adam \cite{kingma_adam} and a starting learning rate of $0.001$. The learning rate is kept constant unless specified in Section~\ref{sec:results}.

\subsection{Similarities and Differences with Existing Techniques}

A first main difference between FLOT and \cite{gu_hplflownet}, \cite{liu_flownet3d}, \cite{wu_pointpwcnet} is the number of parameters which is much smaller for FLOT (see Table~\ref{tab:param_tuning}). Another difference is that we do not use any downsampling and upsampling layers. Unlike \cite{gu_hplflownet}, \cite{wu_pointpwcnet}, we do not use any multiscale analysis to find the correspondences between points. The information between point clouds is mixed only once, as in \cite{liu_flownet3d}, but at the finest sampling resolution and without using skip connections between $g$ and $h$.

We also notice that \cite{gu_hplflownet}, \cite{liu_flownet3d}, \cite{wu_pointpwcnet} rely on a MLP or a convnet applied on the concatenated input features to mix the information between both point clouds. The mixing function is learned and thus not explicit. It is harder to find how the correspondences are effectively done, \ie, identify what input information is kept or not taken into consideration. 
In contrast, the mixing function in FLOT is explicit with only two scalars $\epsilon,\lambda$ adjusted to the training data and whose roles are clearly identified in the OT problem \eqref{eq:optimal_transport_reg}. The core of the OT module is a simple cross-correlation between input features, which is a module easy to interpret, study and visualise. Finally, among all the functions that the convnets/MLPs in \cite{gu_hplflownet}, \cite{liu_flownet3d}, \cite{wu_pointpwcnet} can approximate, it is unlikely that the resulting mixing function actually approximates the Sinkhorn algorithm, or an attention layer, after learning without further guidance than those of the training data.

%% file: sections/results.tex
\subsection{Datasets}

\begin{table}[t]
\centering
\caption{Performance of FLOT on the validation sets of \ftp, \fts, and \fto\ (top). Performance of FLOT measured at the output of the OT module, \ie, before refinement by $h$, on \ftp\ and \fts\ (bottom). The corresponding performance on \fto\ is in the supplementary material. We report average scores and, between parentheses, their standard deviations. Please refer to Section~\ref{sec:study_flot} for more details.}
\label{tab:param_tuning}
\scriptsize
\ra{1.2}
\setlength{\tabcolsep}{2.75pt}
\begin{tabular}{@{} l l l l l l l l l @{}}
\toprule
& Dataset & K & $\epsilon$ & $\lambda / (\lambda + \epsilon)$ & EPE & AS & AR & Out. \\
\midrule
\multirow{13}{*}{\begin{sideways}\bf With flow refinement\end{sideways}} 
& \multirow{4}{*}{\ftp} 
& FLOT$_0$ 
    & $0.03$ \tiny $(0.00)$ 
    & $0$ \tiny (fixed)
    & $0.0026$ \tiny $(0.0005)$ 
    & $99.56$ \tiny $(0.08)$ 
    & $99.69$ \tiny $(0.05)$ 
    & $0.44$ \tiny $(0.10)$ 
    \\
&& 1 
    & $0.03$ \tiny $(0.00)$ 
    & $0.70$ \tiny $(0.00)$
    & $0.0011$ \tiny $(0.0001)$ 
    & $99.83$ \tiny $(0.01)$ 
    & $99.89$ \tiny $(0.01)$ 
    & $0.17$ \tiny $(0.01)$ 
    \\
&& 3 
    & $0.03$ \tiny $(0.00)$ 
    & $0.82$ \tiny $(0.00)$
    & $0.0009$ \tiny $(0.0001)$ 
    & $99.85$ \tiny $(0.01)$ 
    & $99.90$ \tiny $(0.01)$ 
    & $0.16$ \tiny $(0.01)$ 
    \\
&& 5 
    & $0.03$ \tiny $(0.00)$ 
    & $0.88$ \tiny $(0.00)$
    & $0.0009$ \tiny $(0.0001)$ 
    & $99.84$ \tiny $(0.02)$ 
    & $99.90$ \tiny $(0.01)$ 
    & $0.17$ \tiny $(0.02)$ 
    \\
\cmidrule{2-9}
& \multirow{4}{*}{\fts} 
& FLOT$_0\quad$ 
    & $0.03$ \tiny $(0.00)$ 
    & $0$ \tiny (fixed)
    & $0.0811$ \tiny $(0.0005)$ 
    & $50.32$ \tiny $(0.34)$ 
    & $83.08$ \tiny $(0.24)$ 
    & $52.15$ \tiny $(0.34)$
    \\
&& 1       
    & $0.03$ \tiny $(0.00)$ 
    & $0.64$ \tiny $(0.01)$ 
    & $0.0785$ \tiny $(0.0003)$ 
    & $50.91$ \tiny $(0.52)$ 
    & $83.67$ \tiny $(0.10)$ 
    & $51.73$ \tiny $(0.38)$ 
    \\
&& 3       
    & $0.03$ \tiny $(0.00)$ 
    & $0.59$ \tiny $(0.00)$ 
    & $0.0786$ \tiny $(0.0010)$ 
    & $51.06$ \tiny $(0.95)$ 
    & $83.78$ \tiny $(0.35)$ 
    & $51.72$ \tiny $(0.76)$ 
    \\
&& 5       
    & $0.03$ \tiny $(0.00)$ 
    & $0.56$ \tiny $(0.00)$ 
    & $0.0798$ \tiny $(0.0003)$ 
    & $49.77$ \tiny $(0.50)$
    & $83.39$ \tiny $(0.08)$
    & $52.58$ \tiny $(0.31)$ 
    \\
\cmidrule{2-9}
& \multirow{4}{*}{\fto} 
& FLOT$_0$ 
    & $0.03$ \tiny $(0.00)$ 
    & $0$ \tiny (fixed)
    & $0.1834$ \tiny $(0.0018)$ 
    & $21.94$ \tiny $(0.69)$ 
    & $52.79$ \tiny $(0.53)$ 
    & $77.19$ \tiny $(0.43)$ 
    \\
&& 1 
    & $0.03$ \tiny $(0.00)$ 
    & $0.50$ \tiny $(0.01)$ 
    & $0.1798$ \tiny $(0.0009)$ 
    & $22.01$ \tiny $(0.14)$ 
    & $53.39$ \tiny $(0.24)$ 
    & $76.77$ \tiny $(0.16)$ 
    \\
&& 3 
    & $0.03$ \tiny $(0.00)$ 
    & $0.34$ \tiny $(0.00)$ 
    & $0.1797$ \tiny $(0.0014)$ 
    & $22.77$ \tiny $(0.53)$ 
    & $53.74$ \tiny $(0.54)$ 
    & $76.39$ \tiny $(0.43)$ 
    \\
&& 5        
    & $0.03$ \tiny $(0.00)$ 
    & $0.35$ \tiny $(0.01)$ 
    & $0.1813$ \tiny $(0.0020)$ 
    & $22.64$ \tiny $(0.41)$ 
    & $53.58$ \tiny $(0.41)$ 
    & $76.52$ \tiny $(0.46)$ 
    \\
\midrule
\midrule
\multirow{9}{*}{\begin{sideways}\bf No flow refinement\end{sideways}}
&\multirow{4}{*}{\ftp} 
& FLOT$_0$ 
    & \multicolumn{2}{c}{\multirow{4}{*}{\it Same as above}}
    & $0.0026$ \tiny $(0.0006)$ 
    & $99.59$ \tiny $(0.07)$ 
    & $99.70$ \tiny $(0.05)$ 
    & $0.42$ \tiny $(0.10)$ 
    \\
&& 1 &&
    & $0.0010$ \tiny $(0.0001)$ 
    & $99.83$ \tiny $(0.01)$ 
    & $99.89$ \tiny $(0.01)$ 
    & $0.18$ \tiny $(0.01)$ 
    \\
&& 3 &&
    & $0.0009$ \tiny $(0.0000)$ 
    & $99.85$ \tiny $(0.01)$ 
    & $99.90$ \tiny $(0.01)$ 
    & $0.16$ \tiny $(0.01)$ 
    \\
&& 5 &&
    & $0.0010$ \tiny $(0.0001)$ 
    & $99.84$ \tiny $(0.03)$ 
    & $99.90$ \tiny $(0.01)$ 
    & $0.17$ \tiny $(0.02)$ 
    \\
\cmidrule{2-9}
&\multirow{4}{*}{\fts} 
& FLOT$_0$ 
    & \multicolumn{2}{c}{\multirow{4}{*}{\it Same as above}}
    & $0.1789$ \tiny $(0.0004)$ 
    & $17.57$ \tiny $(0.07)$ 
    & $43.34$ \tiny $(0.08)$ 
    & $75.34$ \tiny $(0.07)$
    \\
&& 1 &&      
    & $0.1721$ \tiny $(0.0005)$ 
    & $18.24$ \tiny $(0.11)$ 
    & $44.64$ \tiny $(0.14)$ 
    & $74.54$ \tiny $(0.11)$ 
    \\
&& 3 &&  
    & $0.1764$ \tiny $(0.0003)$ 
    & $17.64$ \tiny $(0.07)$ 
    & $43.52$ \tiny $(0.10)$ 
    & $75.09$ \tiny $(0.07)$ 
    \\
&& 5 &&  
    & $0.1761$ \tiny $(0.0009)$ 
    & $17.68$ \tiny $(0.13)$
    & $43.60$ \tiny $(0.23)$
    & $75.07$ \tiny $(0.13)$ 
    \\
\bottomrule
\end{tabular}
\end{table}

As in related works, we train our network under full supervision using FlyingThings3D \cite{mayer_ft3d} and test it on FlyingThings3D and KITTI Scene Flow \cite{menze_scene_flow_15,menze_scene_flow_18}. However, none of the datasets provide point clouds directly. This information needs to be extracted from the original data. There is at least two slightly different ways of extracting these $3$D data, and we report results for both versions for a better assessment of the performance. The first version of the datasets are prepared\footnote{Code and pretrained model available at \url{https://github.com/laoreja/HPLFlowNet}.} as in \cite{gu_hplflownet}. No occluded point remains in the processed point clouds. We denote these datasets \fts\ and \kittis. The second version of the datasets are the ones prepared\footnote{Code and datasets available at~\url{https://github.com/xingyul/flownet3d}.} by \cite{liu_flownet3d} and denoted \fto\ and \kittio. These datasets contains points where the flow is occluded. These points are present at the input and output of the networks but are not taken into account to compute the training loss \eqref{eq:loss} nor the performance metrics, like in \cite{liu_flownet3d}. Further information about the datasets is in the supplementary material. Note that we keep aside $2000$ examples from the original training sets of \fts\ and \fto\ as validation sets, which are used in Section~\ref{sec:study_flot}.

\subsection{Performance Metrics}

We use the four metrics adopted in \cite{gu_hplflownet}, \cite{liu_flownet3d}, \cite{wu_pointpwcnet}: the end point error $\text{EPE}$; two measures of accuracy, denoted by AS and AR, computed with different thresholds on the $\text{EPE}$; a percentage of outliers also computed using a threshold on the $\text{EPE}$. The definition of these metrics is recalled in the supplementary material.

Let us highlight that the performance reported on \kittis\ and \kittio\ are obtained by using the model trained on \fts\ and \fto, respectively without fine tuning. We do not adapt the model for any of the method. We nevertheless make sure that the $xyz$ axes are in correspondence for all datasets.

\subsection{Study of FLOT}
\label{sec:study_flot}

We use \fts, \fto\ and \ftp\ to check what values the OT parameters $\epsilon, \lambda$ reach after training, to study the effect of $K$ on the FLOT's performance and compare it with that of FLOT$_0$. \ftp\ is exactly the same dataset as \fts\ except that we enforce $\vec{p} + \vec{f} = \ma{P} \vec{q}$ when sampling the point to simulate the perfect world setting. The sole role of this ideal dataset is to confirm that the OT model holds in the perfect world, the starting point of our design.

For these experiments, training is done at $n = 2048$ for $40$ epochs and takes about 9 hours. Each model is trained $3$ times starting from a different random draw of $\theta$ to take into account variations due to initialisation. Evaluation is performed at $n=2048$ on the validation sets. Note that the $n$ points are drawn at random also at validation time. To take into account this variability, validation is performed $5$ different times with different draws of the points for each of the trained model. For each score and model, we thus have access to $15$ values whose mean and standard deviation are reported in Table~\ref{tab:param_tuning}. We present the scores obtained before and after refinement by $h$.

\begin{figure}[t]
\centering
\includegraphics[width=.4\textwidth]{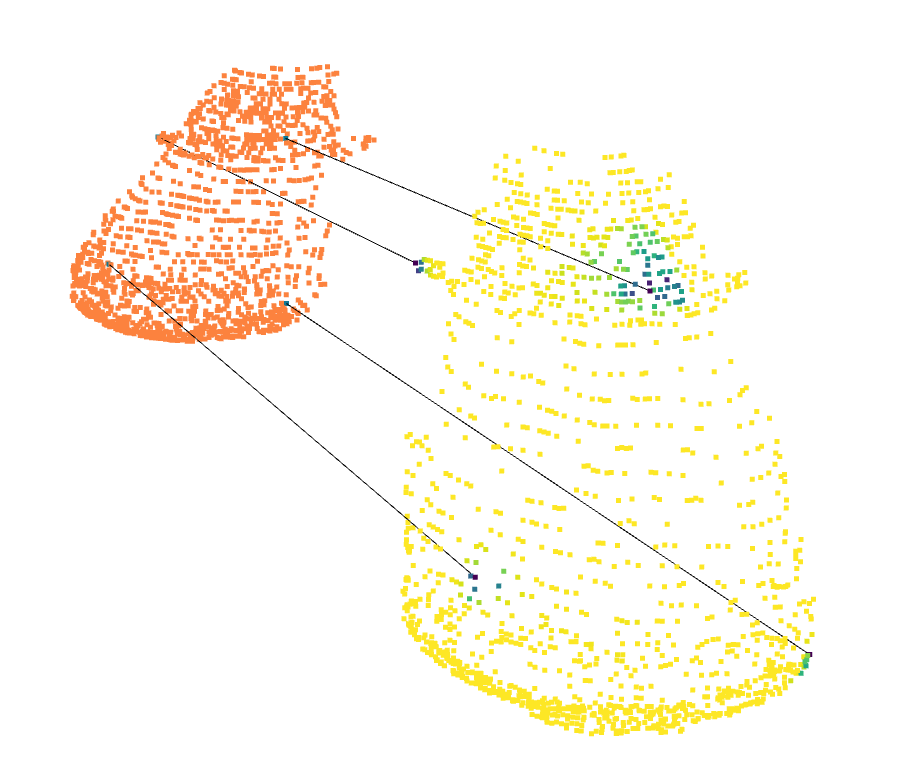}
\includegraphics[width=.4\textwidth]{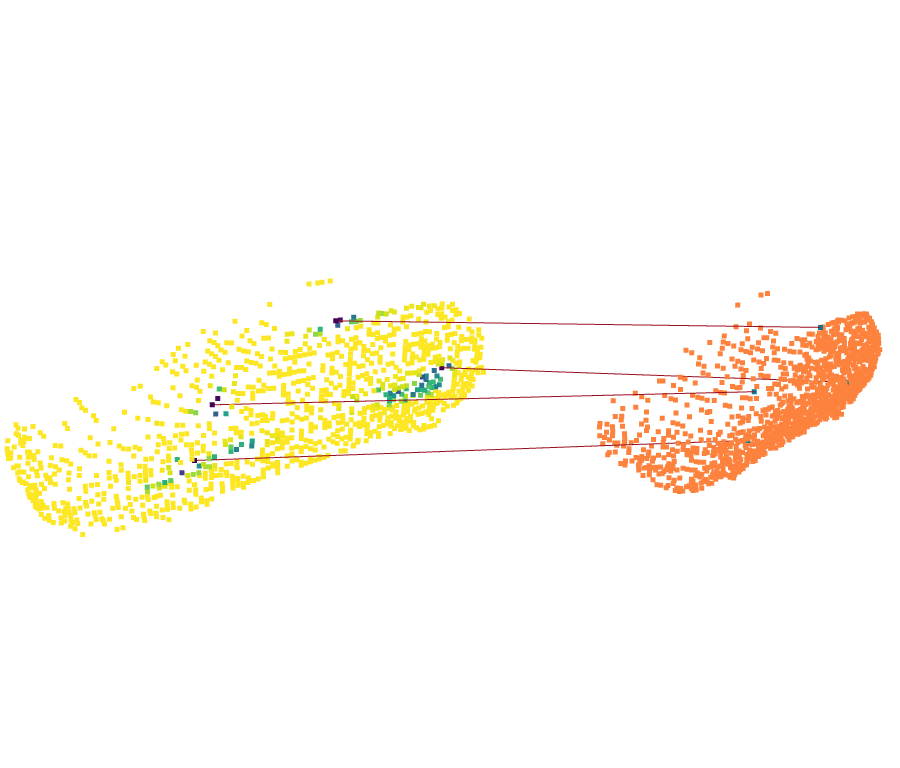}
\caption{Illustration of correspondences, found by FLOT ($K=1$) trained on $n = 8192$ (see Section~\ref{sec:perf_simpler_dataset}),  between $\vec{p}$ and $\vec{q}$ in two different scenes of \kittis. We isolated one car in each of the scenes for better visualisation. The point cloud $\vec{p}$ captured at time $t$ is represented in orange. The lines show the correspondence between a query point $\vec{p}_i$ and the corresponding point $\vec{q}_{j^*}$ in $\vec{q}$ on which most the mass is transported: $j^* = \argmax_j \ma{T}_{ij}$. The colormap on $\vec{q}$ represents the values in $\ma{T}_i$ where yellow corresponds to $0$ and blue indicates the maximum entry in $\ma{T}_i$ and show how the mass is concentrated around $\vec{q}_{j^*}$.}
\label{fig:displacement_cost}
\end{figure}

First, we notice that $\epsilon = 0.03$ for all model after training. We recall that we applied a constant offset of $0.03$ to prevent numerical errors occurring in Alg.~\ref{alg:sinkhorn} in the exponential function when reaching to small value of $\epsilon$. Hence, the entropic regularisation, or, equivalently, the temperature in FLOT$_0$, reaches its smallest possible value. Such small values favour sparse transport plans $\ma{T}$, yielding sparse correspondences between $\vec{p}$ and $\vec{q}$. An illustration of these sparse correspondences is provided in Fig.~\ref{fig:displacement_cost}. We observe that the correspondences are accurate and that the mass is well concentrated around the target points, especially when these points are near corners of the object.

\setlength{\tabcolsep}{11.5pt}
\begin{table}[t]
\centering
\caption{Performance on \fts\ and \kittis. The scores of FlowNet3D and HPLFlowNet are obtained from \cite{gu_hplflownet}. We also report the scores of PointPWC-Net available in \cite{wu_pointpwcnet}, as well as those obtained using the official implementation$^\dagger$. Italic entries are for methods publicly available but not yet published at submission time.}
\label{tab:perf_pure_datasets}
\ra{1.2}
\scriptsize
\begin{tabular}{@{}l l l l l l l @{}}
\toprule
Dataset & Method & EPE & AS & AR & Out. & Size (MB) \\ 
\midrule
\multirow{5}{*}{\fts\ } 
& FlowNet3D \cite{liu_flownet3d} 
    & $0.114$ 
    & $41.2$ 
    & $77.1$ 
    & $60.2$ 
    & $15$ \\
& HPLFlowNet \cite{gu_hplflownet} 
    & $0.080$ 
    & $61.4$ 
    & $85.5$ 
    & $42.9$ 
    & $77$ \\
& FLOT ($K = 1$)
    & $\bf 0.052$ 
    & $\bf 73.2$
    & $\bf 92.7$
    & $\bf 35.7$ 
    & $\bf 0.44$ \\
\cmidrule{2-7}
& \it PointPWC-Net \cite{wu_pointpwcnet}\
    & $\it 0.059$    
    & $\it 73.8$ 
    & $\it 92.8$ 
    & $\it 34.2$ 
    & $\it 30$ \\
& \it PointPWC-Net$^\dagger$
    & $\it 0.055$    
    & $\it 79.0$ 
    & $\it 94.4$ 
    & $\it 29.8$ 
    & $\it 30$ \\
\midrule
\multirow{5}{*}{\kittis\ }
& FlowNet3D \cite{liu_flownet3d}
    & $0.177$ 
    & $37.4$ 
    & $66.8$ 
    & $52.7$ 
    & $15$ \\
& HPLFlowNet \cite{gu_hplflownet} 
    & $0.117$ 
    & $47.8$ 
    & $77.8$ 
    & $41.0$ 
    & $77$ \\
& FLOT ($K = 1$)
    & $\bf 0.056$ 
    & $\bf 75.5$
    & $\bf 90.8$
    & $\bf 24.2$ 
    & $\bf 0.44$ \\
\cmidrule{2-7}
& \it PointPWC-Net \cite{wu_pointpwcnet} 
    & $\it 0.069$ 
    & $\it 72.8$
    & $\it 88.8$
    & $\it 26.5$
    & $\it 30$ \\
& \it PointPWC-Net$^\dagger$ \ 
    & $\it 0.067$ 
    & $\it 78.5$
    & $\it 90.6$
    & $\it 22.8$
    & $\it 30$ \\
\bottomrule
\end{tabular}
\end{table}

Second, the power $\lambda / (\lambda + \epsilon)$, which controls the mass regularisation, reaches higher values on \ftp\ than \fto. This is the expected behaviour as \ftp\ contains no imperfection and \fto\ contains occlusions. The values reached on \fts\ are in between those reached on \ftp\ than \fto. This is also the expected behaviour as \fts\ is free of occlusions and the only imperfections are the different sampling of the scene as $t$ and $t+1$.

Third, on \ftp, FLOT reduces by $2$ the EPE compared to FLOT$_0$, which nevertheless already yields good results. Increasing $K$ from $1$ to $3$ further reduces the error and stabilises at $K=5$. This validates the OT model in our the perfect world setting: the OT optimum and perfect world optimum coincide. 

Fourth, on \fts\ and \fto, the average scores are better for FLOT than FLOT$_0$, except for two metrics at $K=5$ on \fts. The nevertheless good performance of FLOT$_0$ indicates that most of it is due to the trained transport cost $\ma{C}$. On \fts\ and \fto, changing $K$ from $1$ to $3$ has less impact on the EPE than on \ftp. We also detect a slight decrease of performance when increasing $K$ from $3$ to $5$. The OT model \eqref{eq:optimal_transport_reg} can only be an approximate model of the (simulated) real-world. The real-world optimum and OT optimum do not coincide. Increasing $K$ brings us closer to the OT optimum but not necessarily always closer to the real-world optimum. $K$ becomes an hyper-parameter that should be adjusted. In the following experiments, we use $K=1$ or $K=3$. 

Finally, the absence of $h$ has no effect on the performance on \ftp, with FLOT still performing better than FLOT$_0$. This shows that OT module is able to estimate accurately the ideal permutation matrix $\ma{P}$ on its own and that the residual network $h$ is not needed in this ideal setting. However, $h$ plays a important role on the more realistic datasets \fts\ and \fto, with an EPE divided by around $2$ when present.

\subsection{Performance on \fts\ and \kittis}
\label{sec:perf_simpler_dataset}

We compare the performance achieved by FLOT and the alternative methods on \fts\ and \kittis\ in Table~\ref{tab:perf_pure_datasets}. We train FLOT using $n=8192$ points, as in \cite{gu_hplflownet}, \cite{wu_pointpwcnet}. The learning rate is set to $0.001$ for $50$ epochs before dividing it by $10$ and continue training for $10$ more epochs.

The scores of FlowNet3D and HPLFlowNet are obtained directly from \cite{gu_hplflownet}. We report the scores of PointPWC-net available in \cite{wu_pointpwcnet}, as well as the better scores we obtained using the associated code and pretrained model.\footnote{Code and pretrained model available at \url{https://github.com/DylanWusee/PointPWC}.} The model sizes are obtained from the supplementary material of \cite{liu_flownet3d} for FlowNet3D, and from the pretrained models provided by \cite{gu_hplflownet} and \cite{wu_pointpwcnet}. HPLFlowNet, PointPWC-Net and FLOT contain $19$ M, $7.7$ M, and $0.11$ M parameters, respectively.

FLOT performs better than FlowNet3D and HPLFlowNet on both \fts\ and \kittis. FLOT achieves a slightly better EPE than PointPWC-Net on \kittis\ and a similar one on \fts. However, PointPWC-Net achieves better accuracy and has less outliers. FLOT is the method that uses the less trainable parameters ($69$ times less than PointPWC-Net).

\begin{figure}[t]
\centering
\begin{minipage}{.49\textwidth}
\centering
\scriptsize
Inputs $\vec{p}$ (orange) and $\vec{q}$ (blue)
\includegraphics[width=\textwidth]{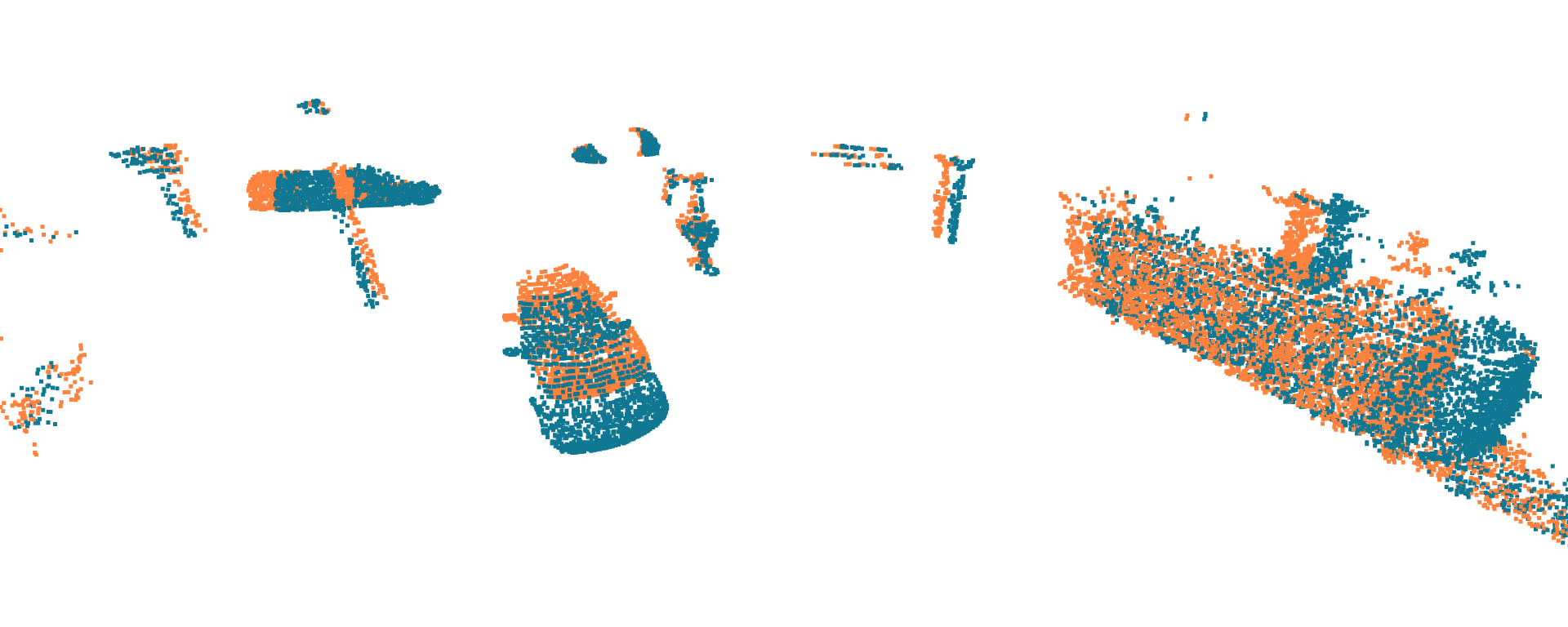}
\end{minipage}
\begin{minipage}{.49\textwidth}
\centering
\scriptsize
Ground truth $\vec{p} + \vec{f}$ (orange) and input $\vec{q}$ (blue)
\includegraphics[width=\textwidth]{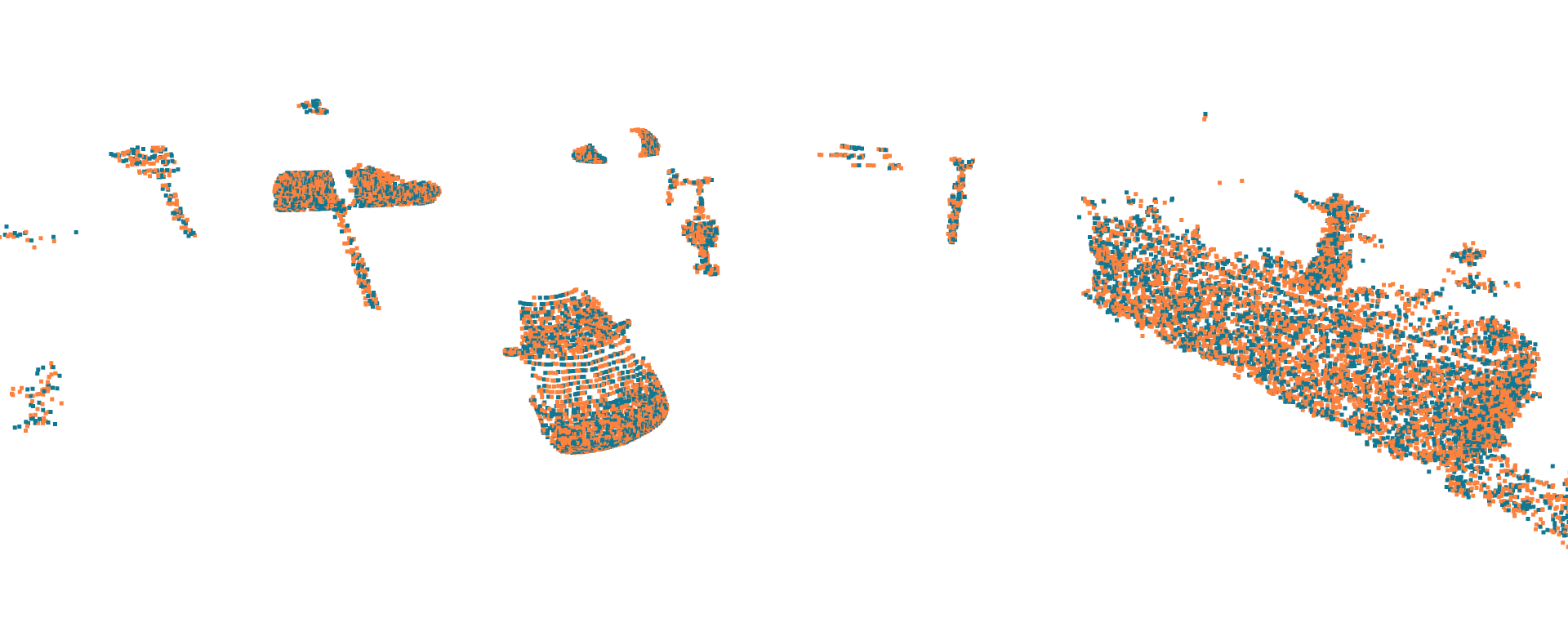}
\end{minipage}
\\
\begin{minipage}{.49\textwidth}
\centering
\scriptsize
Estimated $\vec{p} + \tilde{\vec{f}}$ (orange) and input $\vec{q}$ (blue)
\includegraphics[width=\textwidth]{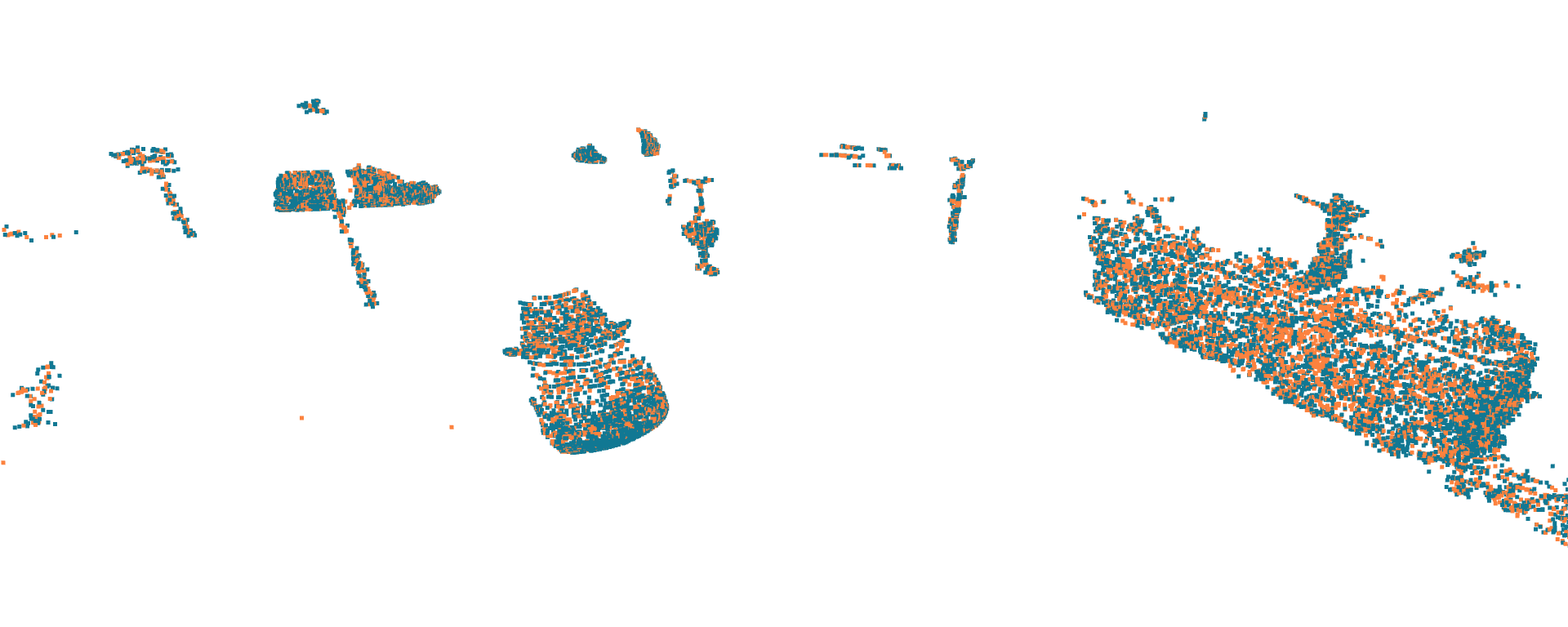}
\end{minipage}
\begin{minipage}{.49\textwidth}
\centering
\scriptsize
Refined $\vec{p} + \vec{f}_{\rm est}$ (orange) and input $\vec{q}$ (blue)
\includegraphics[width=\textwidth]{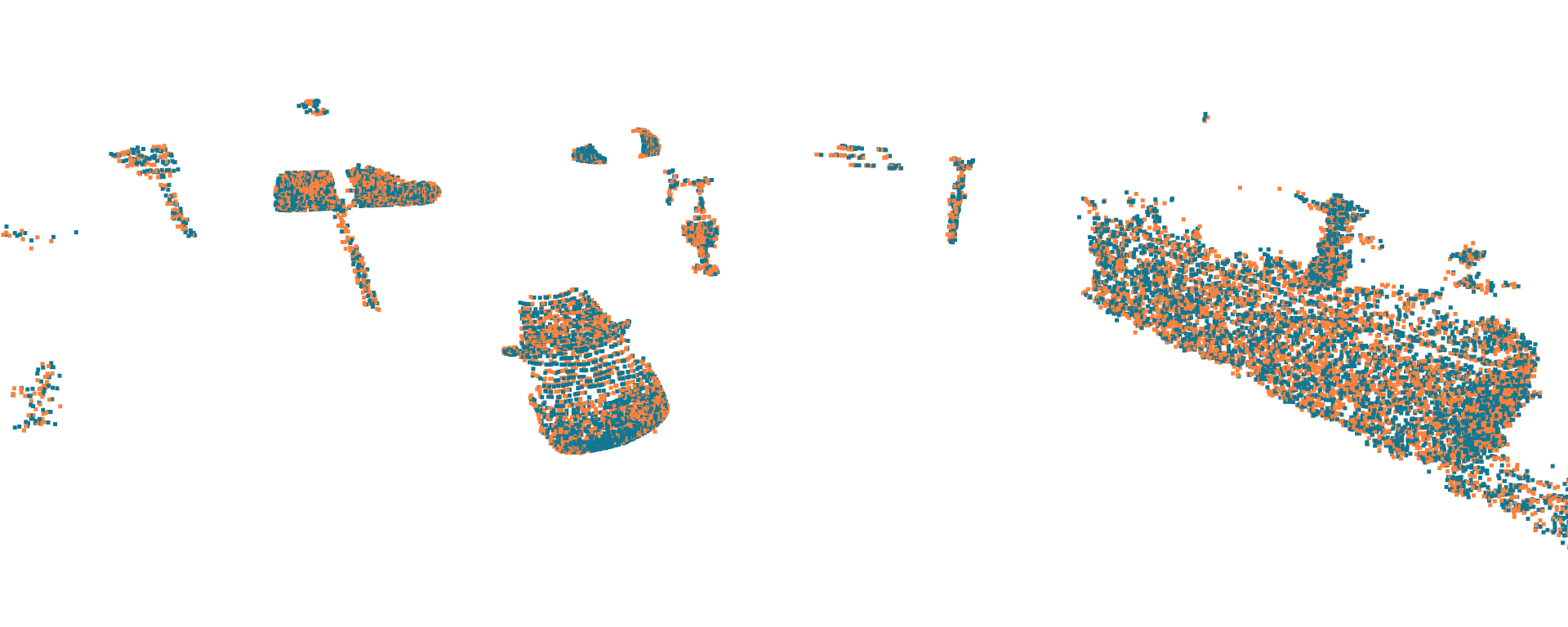}
\end{minipage}
\hrule
\textit{}\\
\begin{minipage}{.49\textwidth}
\centering
\scriptsize
Inputs $\vec{p}$ (orange) and $\vec{q}$ (blue)
\includegraphics[width=\textwidth]{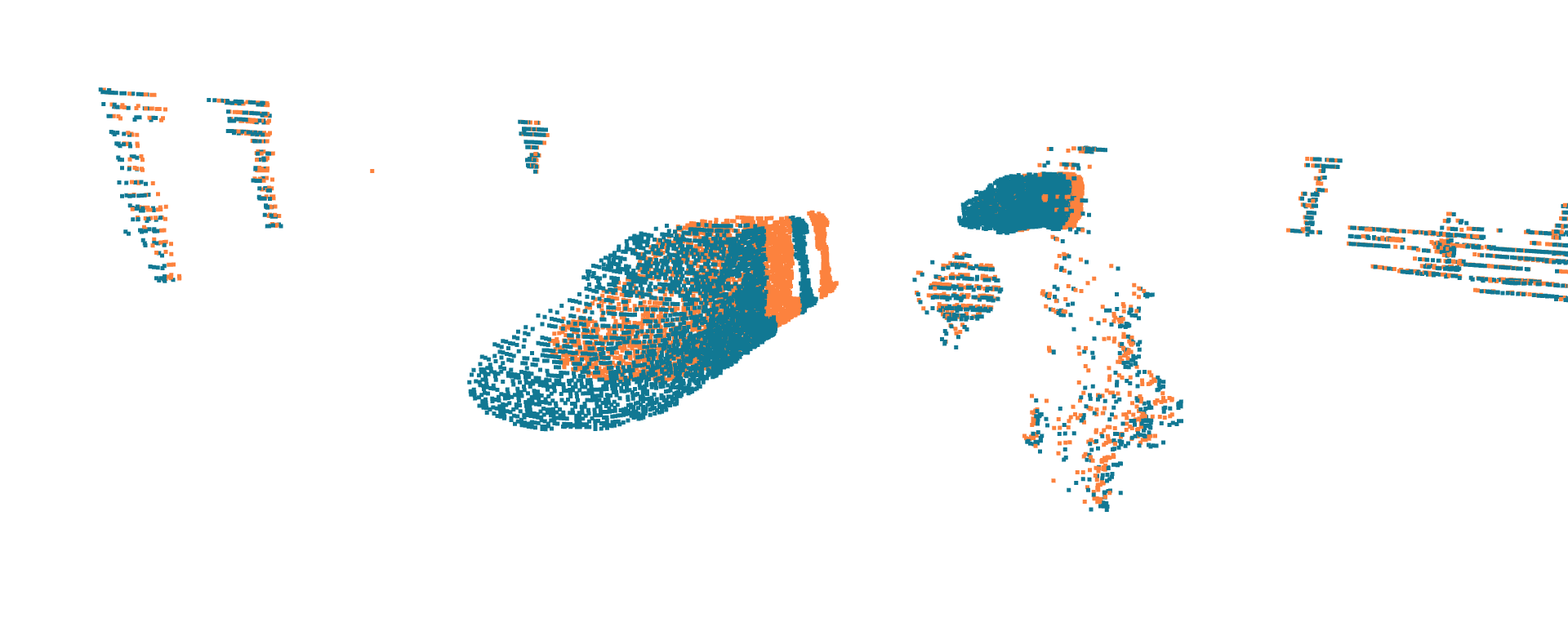}
\end{minipage}
\begin{minipage}{.49\textwidth}
\centering
\scriptsize
Ground truth $\vec{p} + \vec{f}$ (orange) and input $\vec{q}$ (blue)
\includegraphics[width=\textwidth]{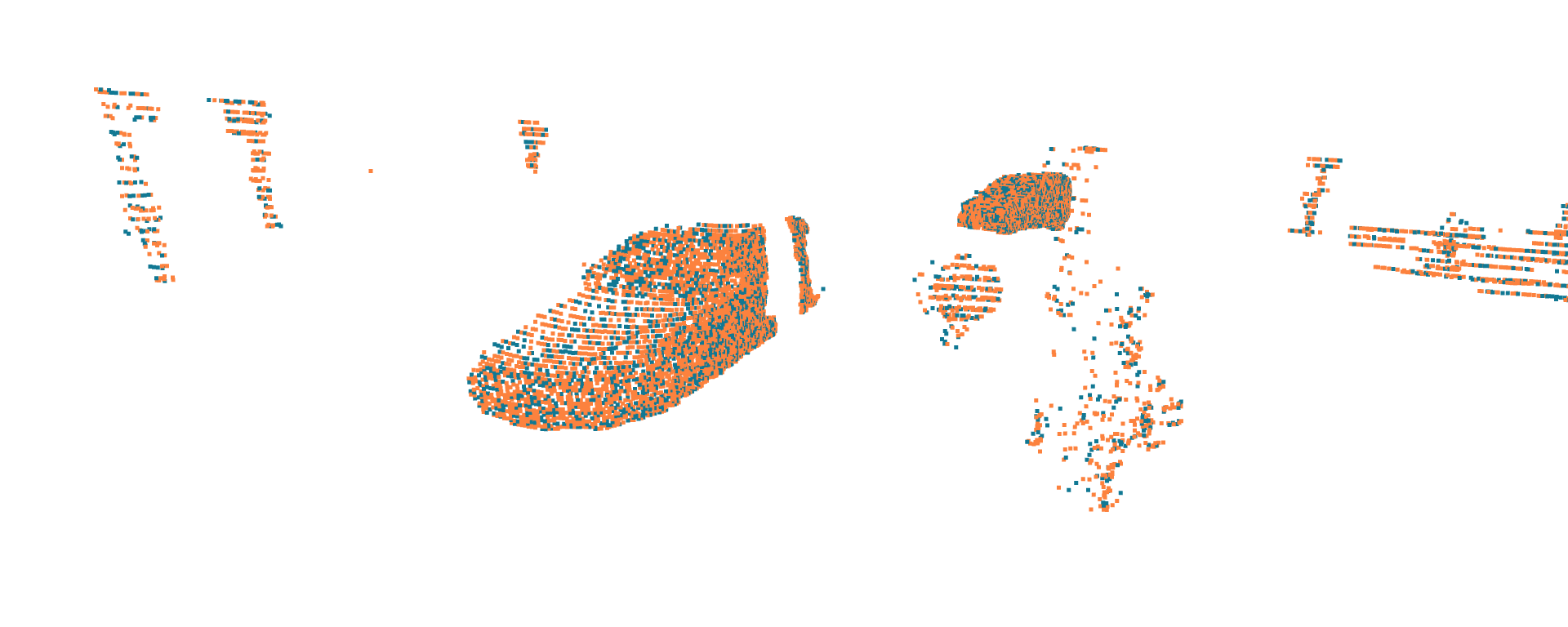}
\end{minipage}
\\
\begin{minipage}{.49\textwidth}
\centering
\scriptsize
Estimated $\vec{p} + \tilde{\vec{f}}$ (orange) and input $\vec{q}$ (blue)
\includegraphics[width=\textwidth]{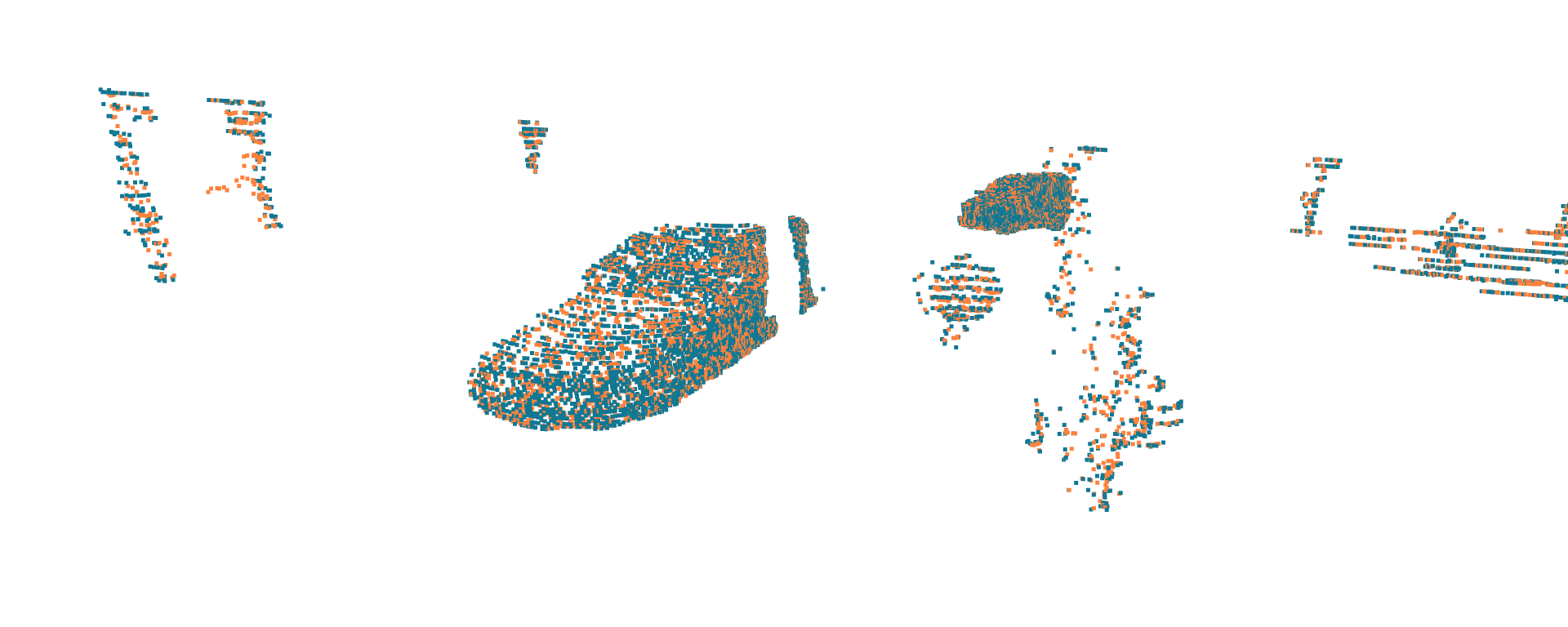}
\end{minipage}
\begin{minipage}{.49\textwidth}
\centering
\scriptsize
Refined $\vec{p} + \vec{f}_{\rm est}$ (orange) and input $\vec{q}$ (blue)
\includegraphics[width=\textwidth]{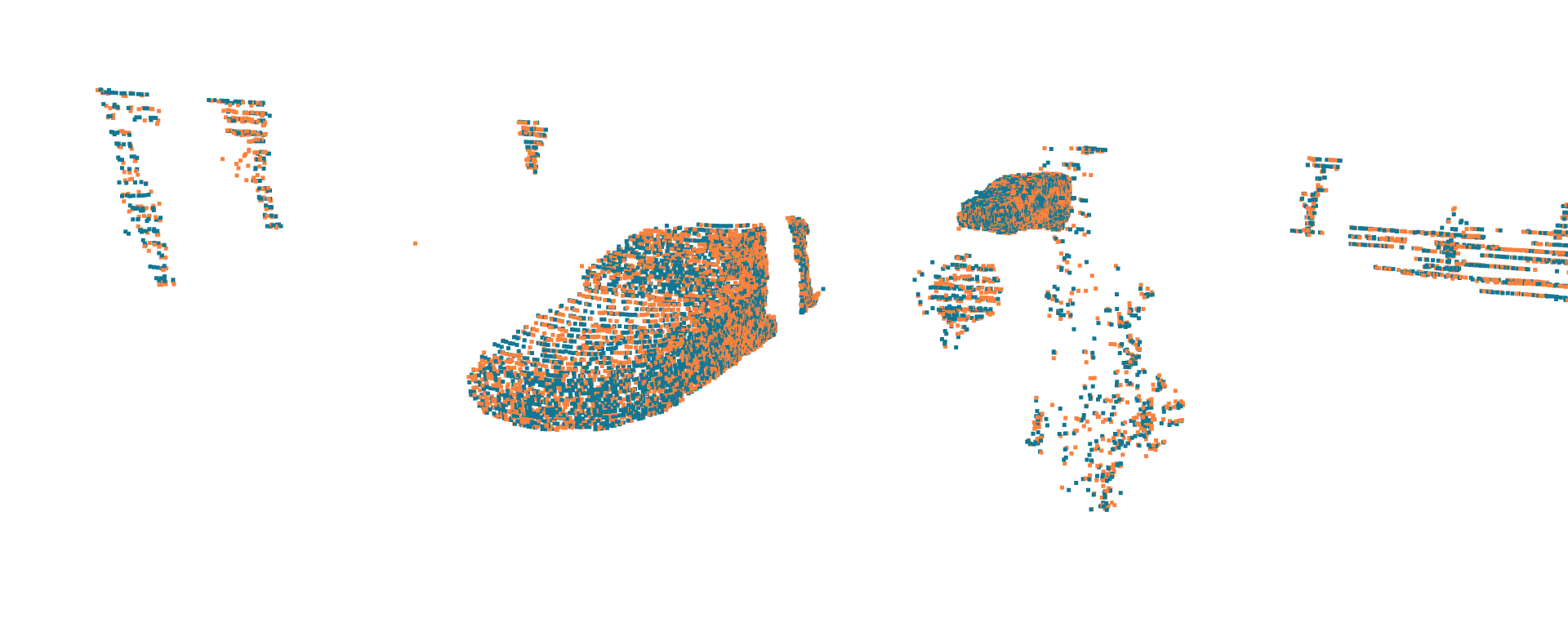}
\end{minipage}
\caption{Two scene from \kittis\ with input point clouds $\vec{p}$, $\vec{q}$ along with the ground truth $\vec{p} + \vec{f}$, estimated $\vec{p} + \tilde{\vec{f}}$ and refined $\vec{p} + \vec{f}_{\rm est}$ using FLOT ($K = 1$) at $n = 8192$.}
\label{fig:features}
\end{figure}

We illustrate in Fig.~\ref{fig:features} the quality of the scene flow estimation for two scenes of \kittis. We notice that FLOT aligns correctly all the objects. We also remark that the flow $\tilde{f}$ estimated at the output of the OT module is already of good quality, even though the performance scores are improved after refinement.

\subsection{Performance on \fto\ and \kittio}

\setlength{\tabcolsep}{19pt}
\begin{table}[t]
\centering
\caption{Performance on \fto\ and \kittio.}
\label{tab:perf_occ_datasets}
\ra{1.2}
\scriptsize
\begin{tabular}{@{}l l l l l l @{}}
\toprule
Dataset & Method & EPE & AS & AR & Out.\\ \midrule
\multirow{4}{*}{\fto}  
& FlowNet3D \cite{liu_flownet3d}
    & $0.160$ 
    & $25.4$ 
    & $58.5$ 
    & $78.9$ \\
& FLOT${}_0$
    & $0.160$ 
    & $33.8$ 
    & $63.8$ 
    & $70.5$ \\
& FLOT ($K=1$)
    & $\bf 0.156$
    & $\bf 34.3$
    & $\bf 64.3$
    & $\bf 70.0$ \\
& FLOT ($K=3$)
    & $0.161$ 
    & $32.3$ 
    & $62.7$ 
    & $71.7$ \\
\midrule
\multirow{4}{*}{\kittio\ }
& FlowNet3D \cite{liu_flownet3d}
    & $0.173$ 
    & $27.6$ 
    & $60.9$ 
    & $64.9$ \\
& FLOT${}_0$
    & $\bf 0.106$ 
    & $\bf 45.3$ 
    & $73.7$ 
    & $46.7$\\
& FLOT ($K=1$)
    & $0.110$ 
    & $41.9$ 
    & $72.1$ 
    & $48.6$ \\
& FLOT ($K=3$)
    & $0.107$ 
    & $45.1$ 
    & $\bf 74.0$ 
    & $\bf 46.3$ \\
\bottomrule
\end{tabular}
\end{table}

We present another comparison between FlowNet3D and FLOT using \fto\ and \kittio, originally used in \cite{liu_flownet3d}. We train FlowNet3D using the associated official implementation. We train FLOT and FLOT$_0$ on $n=2048$ points using a learning rate of $0.001$ for $340$ epochs before dividing it by $10$ and continue training for $60$ more epochs.

The performance of both methods is reported in Table~\ref{tab:perf_occ_datasets}. We notice that FLOT and FLOT$_0$ achieve a better accuracy than FlowNet3D with an improvement of AS of $8.8$ points on \fto\ and $17.7$ on \kittio. The numbers of outliers are reduced by the same amount. FLOT at $K = 1$ performs the best with FLOT$_0$ close behind. On \kittio, the best performing model are those of FLOT$_0$ and FLOT at $K = 3$.

The reader can remark that the results of FlowNet3D are similar to those reported in \cite{liu_flownet3d} but worse on \kittio. The evaluation on \kittio\ is done differently in \cite{liu_flownet3d}: the scene is divided into chunks and the scene flow is estimated within each chunk before a global aggregation. In the present work, we keep the evaluation method consistent with that of Section~\ref{sec:perf_simpler_dataset} by following the same procedure as in \cite{gu_hplflownet}, \cite{wu_pointpwcnet}: the trained model is evaluated by processing the full scene in one pass using $n$ random points from the scene.

%

%% file: sections/conclusion.tex
We proposed and studied a method for scene flow estimation built using optimal transport tools. It can achieves similar performance to that of the best performing method while requiring much less parameters. We also showed that the learned transport cost is responsible for most of the performance. This yields a simpler method FLOT$_0$, which performs nearly as well as FLOT.

We also noticed that the presence of occlusions affects the performance of FLOT negatively. The proposed relaxation of the mass constraints in Eq.~\eqref{eq:optimal_transport_reg} permits us to limit the impact of these occlusions on the performance but does not handle them explicitly. There is thus room for improvements by detecting, \eg, by analysing the effective transported mass, and treating occlusions explicitly.

%% file: sections/appendix.tex
%
\section{Networks architecture}
\label{sec:conv}

\setlength{\tabcolsep}{17pt}
\begin{table}[t]
\begin{center}
\caption{Architecture of $g$ and $h$ where layer $4^{(*)}$ is linear and used in $h$ only.}
\label{tab:architecture}
\ra{1.1}
\begin{tabular}{@{}l l l l l@{}}
Layer $\ell$ 
    & $1$ 
    & $2$ 
    & $3$ 
    & $4^{(*)}$ \\ 
\midrule
MLP size
    & $32$ - $32$ - $32$ 
    & $64$ - $64$ - $64$
    & $128$ - $128$ - $128$
    & $3$
\end{tabular}
\end{center}
\end{table}

The convolutions used in $g$ and $h$ are based on PointNet++ \cite{qi_pointnet} in our implementation. Each convolution layer takes as inputs the point cloud $\vec{r} \in \Rbb^{n \times 3}$ on which the convolution are performed and the features $\vec{\phi}_i^{(\ell)} \in \Rbb^{c'}$, $i = 1, \ldots, n$, coming from the previous layer $\ell$. Note that these features are simply the point coordinates $\vec{r}$ at the input of $g$ and the estimated flow $\tilde{\vec{f}}$ at the input of $h$. For each point $\vec{r}_i$, the indices $\set{N}(\vec{r}_i)$ of the $m = 32$ nearest neighbors to $\vec{r}_i$ in $\vec{r}$ are then computed to obtain $m$ features at point $\vec{r}_i$, each one satisfying
\begin{align}
\left({\vec{\phi}_j^{(\ell)}}^{\adjoint}, \; \vec{r}_j^\adjoint - \vec{r}_i^\adjoint\right)^\adjoint \in \Rbb^{c' + 3},
\end{align}
$j \in \set{N}(\vec{r}_i)$. These features are passed through a $ {\rm MLP}: \Rbb^{c' + 3} \rightarrow \Rbb^{c''}$ consisting of a series of fully connected layer, instance normalisation layer with affine correction \cite{ulyanov_instance}, and leaky ReLu with a negative slope of $0.1$, repeated $3$ times in the same order. Finally, the new feature at point $\vec{r}_i$ is obtained after passing through a final max pooling layer:
\begin{align}
\vec{\phi}_i^{(\ell + 1)} = \max_{j \in \set{N}(\vec{p_i})} \left\{ {\rm MLP} \left[ ({\vec{\phi}_j^{(\ell)}}^{\adjoint}, \; \vec{r}_j^\adjoint - \vec{r}_i^\adjoint)^\adjoint \right] \right\} \in \Rbb^{c''},
\end{align}
where the $\max$ is computed independently for each of the $c''$ channels. These computations are repeated for each point $\vec{r}_i$ of the point cloud using the same MLP. The networks $g$ and $h$ share the same architecture, which is given in Table \ref{tab:architecture}. Note nevertheless that the weights are not shared between $g$ and $h$.

%
\section{Datasets}

The datasets  \fts\ and \kittis\ are prepared\footnote{Code available at \url{https://github.com/laoreja/HPLFlowNet}.} as in \cite{gu_hplflownet}. No occluded point remains in the processed point clouds: one can always find a point $\vec{q}_j$ in $\vec{q}$ such that $\vec{q}_j = \vec{p}_i + \vec{f}_i$ at full sampling rate $N$. However, in practice, most of the points $\vec{p}_i$ do not have a direct matching in $\vec{q}$ as both point clouds are randomly and independently sub-sampled to keep only $n \ll N$ points. This simulates different sampling of the scene. Nevertheless, no object appears or disappears because of occlusions between $t$ and $t+1$. \fts\ contains $19,640$ training examples, from which we keep $2,000$ aside for validation, and $3,824$ test examples. \kittis\ contains $200$ examples for which $142$ are used for test, as in \cite{gu_hplflownet}. We do not use the remaining KITTI examples. The ground points in \kittis\ are removed using a threshold on the height. All points whose depth is larger than $35$ m are removed in both datasets.

The datasets \fto\ and \kittio\ are the prepared\footnote{Datasets available at~\url{https://github.com/xingyul/flownet3d}.} by \cite{liu_flownet3d}. In \fto, masks where the flow is non valid, \eg, due to occlusions, are provided in used in the training loss, like in \cite{liu_flownet3d}. These masks are also used to compute the scores only on valid points at test time for all methods. However, the points where the corresponding flow is non-valid are present at the input of all networks. No mask is provided for \kittio. \fto\ contains $19,999$ training examples, from which we keep $2,000$ aside for validation, and $2,003$ test examples.\footnote{We removed $8$ examples with all points marked as occluded ($7$ in the training set and $4$ in the test set). One example which contains a non valid value in the training dataset is also removed.} \kittio\ contains $150$ test examples. The ground points in \kittio\ are removed by \cite{liu_flownet3d}. All points whose depth is larger than $35$ m are removed in both datasets.

%
\section{Performance metrics}

We use the following four metrics adopted in \cite{gu_hplflownet}, \cite{liu_flownet3d}, \cite{wu_pointpwcnet}:
\begin{itemize}
\item $\text{EPE}_i = \norm{(\vec{f_{\rm est}})_i - \vec{f}_i}_2$: end point error, averaged over all $i$;
\item AS: percentage of points such that $\text{EPE}_i < 0.05$ or $\text{EPE}_i / \norm{\vec{f}_i}_2 < 0.05$;
\item AR: percentage of points such that $\text{EPE}_i < 0.1$ or $\text{EPE}_i / \norm{\vec{f}_i}_2 < 0.1$
\item Out.: percentage of points such that $\text{EPE}_i > 0.3$ or $\text{EPE}_i / \norm{\vec{f}_i}_2 > 0.1$.
\end{itemize}

The above metrics are computed as follows. The point clouds $\vec{p}, \vec{q}$ are obtained by selecting $n$ random points out of the $N$ provided points in the datasets. The flow is estimated and compared to the ground truth flow $\vec{f}$ on these $n$ selected points. The scores are averaged over the whole validation/test set.

%
\section{Additional experimental results}

\subsection{Study of FLOT}

We report in Table \ref{tab:no_refinement} the performance of FLOT obtained at the output of the OT module on \fto. The corresponding performance with refinement are available in the core of the paper. As on \fts, we remark that the refinement permits to improve the EPE by around $2$, confirming its utility in presence of occlusions.

\setlength{\tabcolsep}{12pt}
\begin{table}[t]
\begin{center}
\caption{Performance of FLOT measured at the output of the OT module, \ie, before refinement by $h$, on \fto. We report the average scores and their standard deviations between parentheses.}
\label{tab:no_refinement}
\scriptsize
\ra{1.3}
\begin{tabular}{@{}l l l l l l @{}}
\toprule
Dataset & K & EPE & AS & AR & Out. \\ 
\midrule
\multirow{4}{*}{\fto\ } 
& FLOT$_0$ 
    & $0.3539$ \tiny $(0.0028)$ 
    & $6.98$ \tiny $(0.11)$ 
    & $22.05$ \tiny $(0.28)$ 
    & $88.76$ \tiny $(0.14)$ 
    \\
& 1 
    & $0.3412$ \tiny $(0.0042)$ 
    & $7.55$ \tiny $(0.17)$ 
    & $23.50$ \tiny $(0.40)$ 
    & $88.02$ \tiny $(0.22)$ 
    \\
& 3 
    & $0.3426$ \tiny $(0.0028)$ 
    & $7.38$ \tiny $(0.04)$ 
    & $23.09$ \tiny $(0.05)$ 
    & $88.21$ \tiny $(0.03)$ 
    \\
& 5        
    & $0.3440$ \tiny $(0.0021)$ 
    & $7.32$ \tiny $(0.05)$ 
    & $22.94$ \tiny $(0.16)$ 
    & $88.34$ \tiny $(0.09)$ 
    \\
\bottomrule
\end{tabular}
\end{center}
\end{table}
%

\subsection{Computation time in the OT module}

At $n=2048$, the computation time\footnote{Computed on a Nvidia GeForce RTX 2080 Ti.} in the OT module is $1.4$, $2.0$ and $2.2$ ms for FLOT$_0$, FLOT $K=1$, FLOT $K=3$, respectively. At $n=8192$, the computation time in the OT module is $13.1$, $16.0$, $17.9$ ms for FLOT$_0$, FLOT $K=1$, FLOT $K=3$, respectively. This represents at most $8\%$ of the total computation time which is itself at most of $27.8$ ms at $n=2048$ and $346$ ms at $n=8192$. Most of the time, at least $67\%$ at $n=2048$ and $86\%$ at $n=8192$, is spent in the feature extractor $g$. This shows that the OT module is responsible for just a small fraction of the total computation time.

Note that the time spent in the OT module is independent of the type of convolution used. Replacing our implementation of PointNet++ with a faster one or choosing a faster convolution will directly improve the computation time spend in $g$ and $h$. Our implementation of the OT module can also be made faster by avoiding to compute densely the cost matrix $\ma{C}$ by restricting the computation to points that are less than $d_{\rm max}$ meters apart, as these points never contribute to $\ma{T}$.